# An Experimental Platform for Multi-spacecraft Phase-Array Communications

Aaditya Ravindran
Space and Terrestrial Robotic Exploration Laboratory
Arizona State University
Tempe, United States

Ravi Teja Nallapu
Space and Terrestrial Robotic Exploration Laboratory
Arizona State University
Tempe, United States

Andrew Warren
Space and Terrestrial Robotic Exploration Laboratory
Arizona State University
Tempe, United States

Alessandra Babuscia
Jet Propulsion Laboratory
California Institute of Technology
Pasadena, California
Alessandra.babuscia@jpl.nasa.gov

Jose Vazco
Jet Propulsion Laboratory
California Institute of Technology
Pasadena, California
Jose.E.Velazco@jpl.nasa.gov

Jekan Thangavelautham
Space and Terrestrial Robotic Exploration Laboratory
Arizona State University
Tempe, Arizona
jekan@asu.edu

*Abstract*— **The emergence of small satellites and CubeSats for interplanetary exploration will mean hundreds if not thousands of spacecraft exploring every corner of the solar-system. Current methods for communication and tracking of deep space probes use ground based systems such as the Deep Space Network (DSN). However, the increased communication demand will require radically new methods to ease communication congestion. Networks of communication relay satellites located at strategic locations such as geostationary orbit and Lagrange points are potential solutions. Instead of one large communication relay satellite, we could have scores of small satellites that utilize phase arrays to effectively operate as one large satellite. Excess payload capacity on rockets can be used to warehouse more small satellites in the communication network. The advantage of this network is that even if one or a few of the satellites are damaged or destroyed, the network still operates but with degraded performance. The satellite network would operate in a distributed architecture and some satellites maybe dynamically repurposed to split and communicate with multiple targets at once. The potential for this alternate communication architecture is significant, but this requires development of satellite formation flying and networking technologies. Our research has found neural-network control approaches such as the Artificial Neural Tissue can be effectively used to control multirobot/multi-spacecraft systems and can produce human competitive controllers. We have been developing a laboratory experiment platform called Athena to develop critical spacecraft control algorithms and cognitive communication methods. We briefly report on the development of the platform and our plans to gain insight into communication phase arrays for space.**

*Keywords—phased array; control; neural networks; testbed*

I. INTRODUCTION

The increasing demand for in-space communications and tracking will require development of new systems to handle the traffic. Space-based communication relay satellites positioned at critical locations such as geostationary orbit and Lagrange points can augment and enhance our current deep space communication capabilities to reach deeper into space and with increased data rates. Our work has focused on using phased arrays [1] of small antennas that work together and act as a single, big antenna for both transmitting and receiving data (Fig. 1). The signals emitted from each small antenna undergo constructive and destructive interference that steers the overall strength and direction of the beam. The relative position and direction of the individual small antennas are critical towards achieving the overall beam shape and direction.

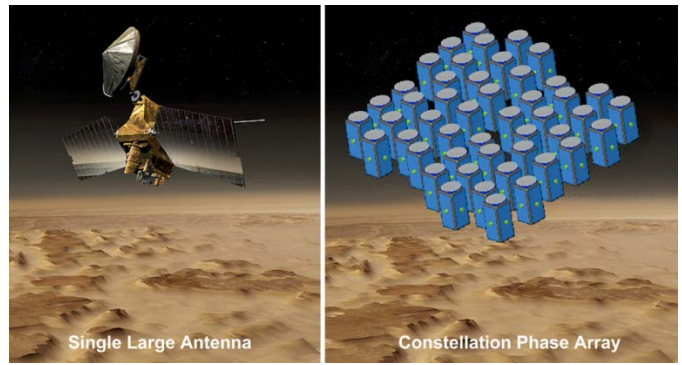

Fig. 1. A network of small-satellites can position their antennas to form large phased array antenna and exceed the capabilities of single, large communication relay satellites.

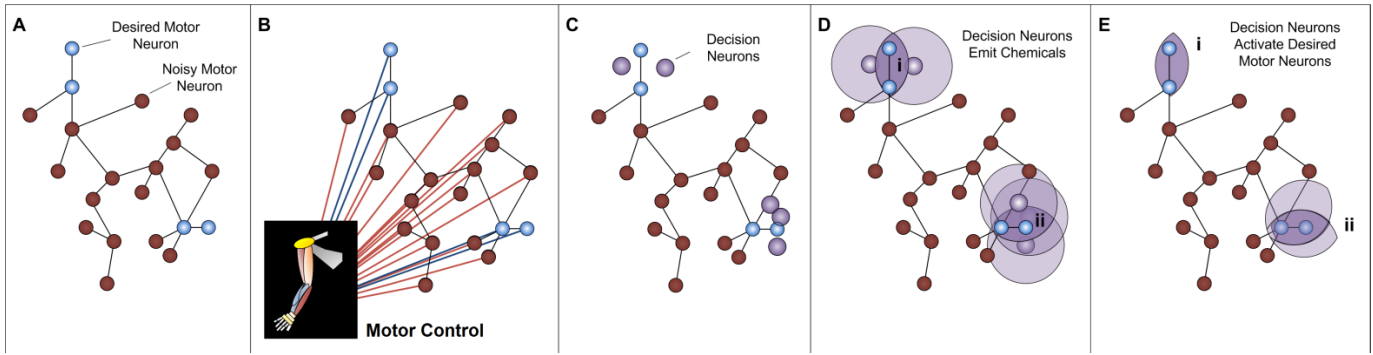

Fig. 2. In a randomly generated tissue, most motor neurons would produce spurious/incoherent output (a) that would `drown out' signals from a few desired motor neurons due to spatial crosstalk [2] (b). This can make training intractable for difficult tasks. Neurotransmitter (chemicals) emitted by decision neurons (c) selectively activate networks of desired motor neurons in shaded regions (i) and (ii) by coarse-coding overlapping diffusion fields as shown (d). This inhibits noisy motor neurons and eliminates spatial crosstalk (e).

Having each antenna on a small spacecraft, we convert this problem into a multi-spacecraft coordination and control problem. Past work has shown decentralized multirobot control can be achieved with each robot utilizing only local sensing and actuation [3],[4],[5],[6],[7]. The challenge is determining what individual behaviors are required to achieve global consensus.

These behaviors maybe developed by hand [3],[4],[5]. This is sufficient when only a few robots are involved. However, as we increase the number of robots, intuitive methods breakdown and we are faced with poor results. As a result, the human designer will then resort to trial and error to further tweak the algorithm. A compelling alternative has been to utilize machine learning to evolve scalable multirobot behaviors using Artificial Neural Tissues (Fig. 2) [7],[8]. The multirobot system is evolved using a fitness function that evaluates the overall performance of the system and undergoes a form of directed trial and error learning utilizing an evolutionary algorithm. A population of individuals competes, so that the fittest individuals live and thrive, while the unfit individuals are culled off. We have found much success utilizing artificial neural networks as the robot controllers. The approach enables the system to find creative behaviors that may not have been thought of by the experimenter [6],[7],[8]. Even more impressive, the controllers can exceed human design controllers for certain multirobot tasks [9].

We are now proceeding to implement this capability on teams of spacecrafts that would utilize phase array technology to effectively generate a large communication antenna in space. The big potential for this technology lies in overall system robustness, immunity to single point failures and for it to be extensible, enabling addition of new arrays to further increase the range and beam strength.

In this short paper, we first present ongoing work in the development of an experiment platform and learning control to coordinate the action of multiple spacecraft to form a large antenna. The controller framework is described in Section II, followed by description of our experiment platform in Section III, formation experiments in Section IV and preliminary conclusions in Section V.

## II. LEARNING CONTROL

The proposed bio-inspired control algorithm for controlling multiple robots or spacecraft is described in this section. It consists of the Artificial Neural Tissue (ANT) architecture [7],[8] and is a developmental program encoded in an artificial `genome,' that constructs a three-dimensional artificial neural network which we call a *neural tissue*. The ANT architecture is evolved using an evolutionary algorithm. The approach enables simultaneous evolution of the topology and contents of the neural tissue. Only a fitness function (a form of goal function) is used to guide the training process. A successful controller needs to perform self-organized task decomposition, taking the goal function, subdividing it and solving the required subtasks to then solve the overall task. The approach has been successfully applied to control of multiple robots to achieve a desired global goal. An example is shown in Fig. 3, where a team of robots learn to excavate given a 3D excavation blueprint [6],[9]. The robots learn to cooperate and efficiently push regolith in unison to form a berm.

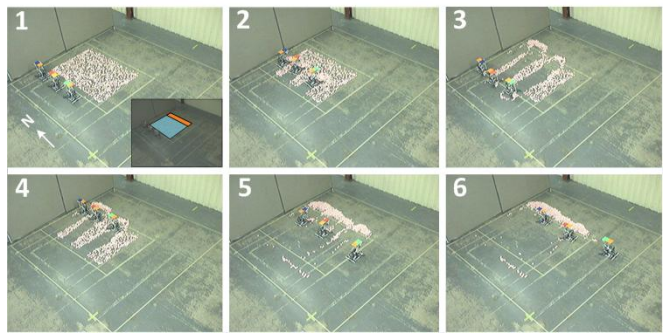

Fig. 3. A team of robots using the ANT controller have learned to cooperate and efficiently perform excavation in formation [6].

The ANT architecture is unique in that it consists of two types of neural units, decision neurons and motor-control neurons, or simply motor neurons. Assume a randomly generated set of motor neurons in a tissue connected electrically by wires (Fig. 2). Chances are most of these neurons will produce incoherent/noisy output, while a few may produce desired functions. If the signal from all of these neurons are summed, then these 'noisy' neurons would drown out the output signal (Fig. 2b) due to spatial crosstalk [2]. Within ANT, decision neurons emit chemicals that diffuse omnidirectionally (shown shaded) (Fig. 2c). By coarse-coding multiple overlapping diffusion fields, the desired motor neurons is selected and ``noisy" neurons inhibited, referred to as neural regulation. With multiple overlapping diffusion fields (Fig. 2d), there is redundancy and when one decision neuron is damaged the desired motor neurons are still selected. A detailed description of the algorithm can be found in [8].

For application to the phase-array formation task, the entire systems of robots would be evaluated by a numerical goal function such as net signal strength or maximum data transfer rate achieved by the array. They would not be given specific commands to form a particular multirobot configuration instead the controllers would need to dynamically learn and determine the best configuration based on the current state of the system. This may include addition of new satellites to the group or removal due to damage or individual reassignment. The algorithm would need to best organize and manage the satellites to achieve the best array configuration for the task at hand. In this approach, training can occur prior to deployment and during operations if needed.

### III. EXPERIMENT PLATFORM

In this section we describe the *Athena* experiment platform that is being developed to test and demonstrate our multispacecraft coordination algorithm to enable phase arrays. Athena is intended to test and characterize the coordination algorithms in a controlled laboratory environment. Promising results would lead to experiments on-orbit using real spacecrafts to demonstrate the concept. This section describes the various modules of the Athena Robots. The robots have 4 operational modules including command and control, flotation, mobility/propulsion and navigation modules.

#### A. Command and Control

The command and control architecture for the Athena robots is shown in Fig. 4. Each robot is equipped with XBEE radio which enables wireless commanding of high-level goals and overrides by a user. The robot controller is autonomous and can be manually programmed or be the product of artificial evolution as described earlier. The controller receives input from the navigation system, while it commands the mobility and flotation system. Two-way communication is performed with the robot control system and for reprogramming of the Software Defined Radios (SDRs).

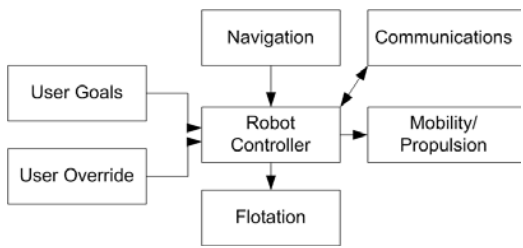

Fig. 4. *Athena* robot control architecture.

#### B. Flotation

The flotation module consists of a gas source which is fed to a series of air-bearings via a solenoid valve. Fig. 5 shows various components of the flotation module. It enables each robot to float much like a hovercraft. Carbon dioxide is used as the gas source on these robots.

The air bearings used have micro-pores on their surface, which shoot out the gas supplied. The gas released from the bearings creates a gas cushion, on the order of a few microns, which makes the bearings, and the structure they support, float on a plane (Fig. 6). It should be noted that the air bearing surfaces are free to swivel, to compensate for irregularities in the surfaces on which they float.

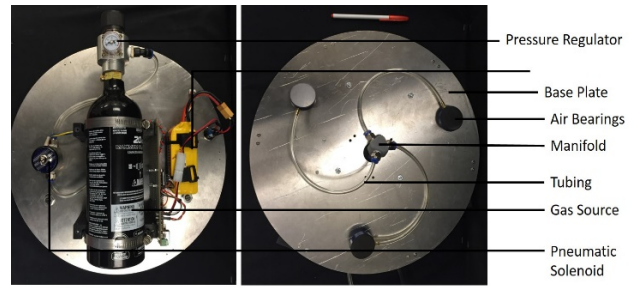

Fig. 5. Top and bottom views of *Athena* robot's flotation unit.

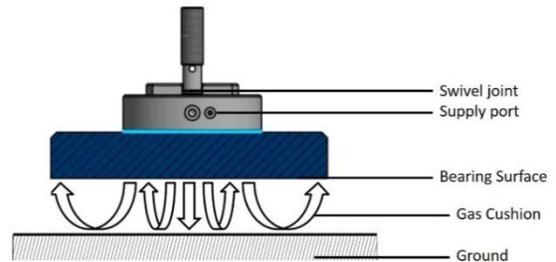

Fig. 6. Air bearing operations.

#### C. Propulsion/Mobility

Once the robot is freely floating, it would use its propulsion system to get from one location to another or achieve a desired orientation. For this experiment system, the propulsion unit consists of 8 ducted fans as shown in Fig. 7. They are used to achieve both translational and rotational movement. On a real spacecraft, the propulsion system would consist of cold-gas, monopropellant or bi-propellant thrusters.

#### D. Navigation

The navigation module is used to provide positon and orientation of the robots. The robots contain an Inertial Measurement Unit (IMU) that includes a gyro and accelerometer to keep track of the orientation. Robot position would be determined using an external localization system that consists of an overhead camera. This overhead localization system would be replaced by a Global Positioning System (GPS) type system in space. In addition, to obtain absolute 'ground truth,' we plan on using a VICON system that provides true position and orientation of the robots.

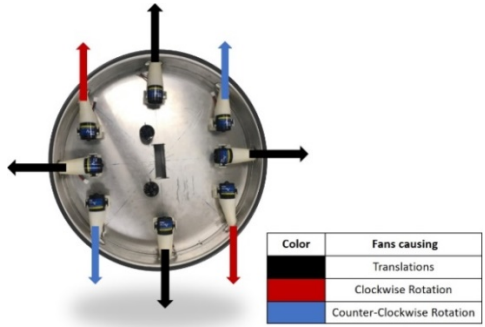

Fig. 7. Propulsion/mobility layout.

*E. Communications*

The communication architecture for *Athena* is shown in Fig. 8. Each robot houses a microcontroller consisting of the Raspberry Pi running GNU Radio which commands the software defined radio, a USRP 205 Mini-i by Ettus.

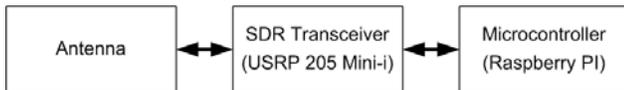

Fig. 8. Communication architecture

Currently, a frequency division multiple access system has been tested on the SDR. The data is transmitted after Gaussian Minimum Shift Keying (GMSK) modulation and a polyphase channelizer to divide the frequency band into several channels. Data is then received through a polyphase synthesizer to map the channels back and demodulated. Some results can be found in Fig. 9. The different peaks in Fig. 9 represent transmission on 4 different channels the frequency band is divided into. Each robot communicates with the group using a preassigned encrypted (128-bit Advanced Encryption Standard) RF channel for formation communication.

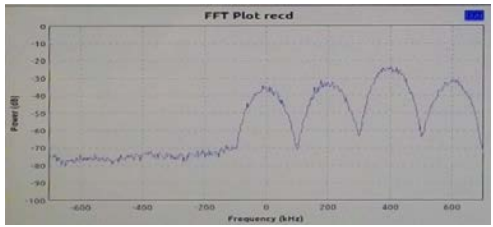

Fig. 9. Receiver frequency response.

IV. EXPERIMENTS

We are performing simple laboratory experiments to demonstrate the phase array concept using 2-5 robots but that can be readily expanded to 100s of units. As noted earlier, it is critical for the robots to assemble into formation (see Fig. 10) and maintain a set distance and relative location to then use the appropriate phase array technique to amplify the overall signal. Manual control of the robot position show promising results, achieving accuracy of ± 1 cm and our efforts are now focused on implementing autonomous control using the ANT controls approach. Once the robots are in formation, they will be faced with simulated conditions such as loss robots and intermittent loss of communication. Under these conditions, the remaining robots will need to dynamically reconfigure themselves and segregate unusable robots to achieve maximum data transmission rates.

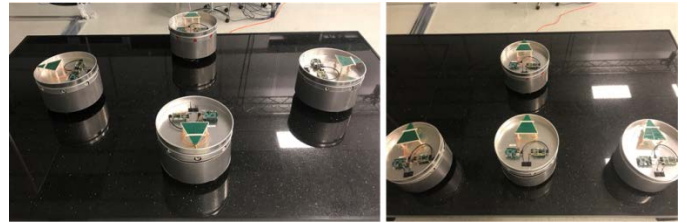

Fig. 10. *Athena* robots in two different formations.

V. CONCLUSION

In this paper we present ongoing development of a laboratory robotics platform called *Athena* that will be used to prototype and characterize multi-spacecraft phase array communication technology in the laboratory. Each robot utilizes air-bearings to float on a granite experiment table and is equipped with a software defined radio An onboard propulsion system utilizing ducted fans enables both attitude control and translational mobility. Our research now focuses on implementing evolvable neural network control technology to enable a team of spacecraft to achieve formation flight with minimal supervision. The intent is for the spacecraft team to adapt to variable conditions including loss or degradation of one or more spacecraft to achieve optimal system performance.